\documentclass[conference]{IEEEtran}
\usepackage{epsfig}
\usepackage{graphicx}
\usepackage{amsmath}
\usepackage{amssymb}
\usepackage{subfig}
\usepackage{dblfloatfix} 
\usepackage{float}
\usepackage{color}
\usepackage[table]{xcolor}
\usepackage[pagebackref=true,breaklinks=true,letterpaper=true,colorlinks,bookmarks=false]{hyperref}
\newcommand{\etal}{\textit{et al.}}

\begin{document}
\title{Object Proposals for Text Extraction in the Wild}

\author{\IEEEauthorblockN{Llu\'is G\'omez and Dimosthenis Karatzas}
\IEEEauthorblockA{Computer Vision Center, Universitat Aut\`onoma de Barcelona\\
Email: \{lgomez,dimos\}@cvc.uab.es}}
\maketitle

\begin{abstract}

Object Proposals is a recent computer vision technique receiving increasing interest from the research community. Its main objective is to generate a relatively small set of bounding box proposals that are most likely to contain objects of interest. The use of Object Proposals techniques in the scene text understanding field is innovative. Motivated by the success of powerful while expensive techniques to recognize words in a holistic way, Object Proposals techniques emerge as an alternative to the traditional text detectors.

In this paper we study to what extent the existing generic Object Proposals methods may be useful for scene text understanding. Also, we propose a new Object Proposals algorithm that is specifically designed for text and compare it with other generic methods in the state of the art. Experiments show that our proposal is superior in its ability of producing good quality word proposals in an efficient way. The source code of our method is made publicly available\footnote{\url{http://github.com/lluisgomez/TextProposals}}.
\end{abstract}


\IEEEpeerreviewmaketitle

\section{Introduction}

Scene Text understanding consists in determining whether a given image contains textual information and if so, localizing it and recognizing its written content. Traditionally this challenging task has been tackled with a multistage pipeline where text detection, extraction, and recognition steps have been treated separately as isolated problems. 
 More recently, an alternative framework has been proposed motivated by the high accuracy of methods for whole word recognition and the emergent use of Object Proposal techniques. This new framework has produced the best performing state-of-the-art methods for scene text end-to-end word spotting \cite{Almazan2015,jaderberg2014reading}.

\begin{figure}[t]
\centering
\includegraphics[width=0.99\linewidth]{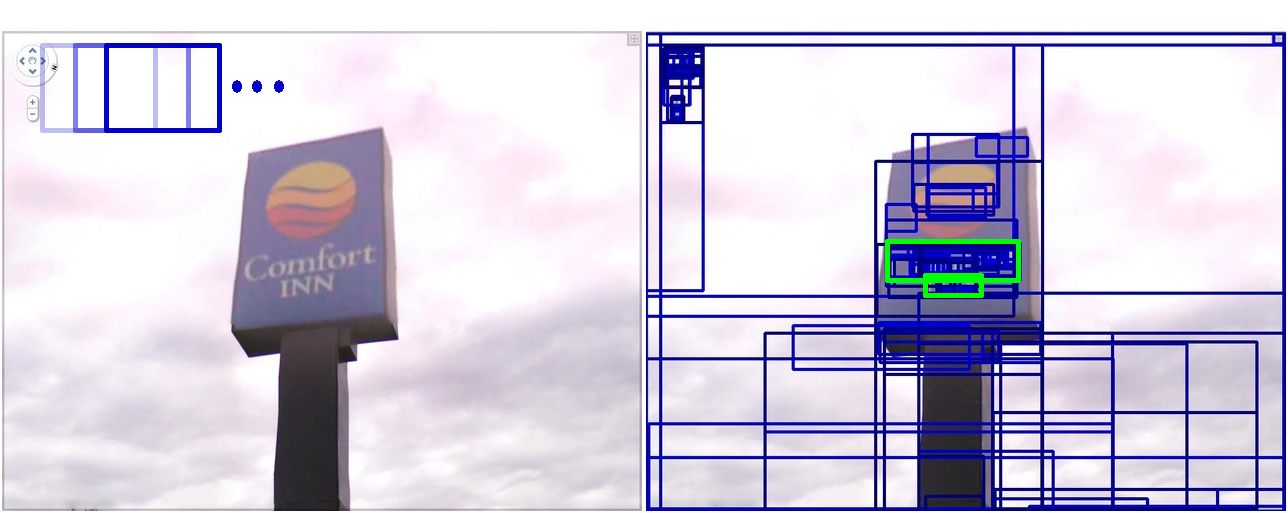}
\caption{Sliding a window for all possible locations, sizes, and aspect ratios represents a considerable waste of resources. The best ranked 250 proposals generated with our text specific selective search method provide 100\% recall and high-quality coverage of words in this particular image.}
\label{fig:250proposals}
\end{figure}

Object Proposals is a recent computer vision technique for generation of high quality object locations. The main interest of such methods is their ability to speed up recognition pipelines that make use of complex and expensive classifiers by considering only a few thousands of bounding boxes. 
 It therefore constitutes an alternative to exhaustive search, which has many well known drawbacks, and enables the efficient use of more powerful classifiers by greatly reducing the search space as shown in Figure~\ref{fig:250proposals}.

In the context of scene text understanding, whole-word recognition methods~\cite{goel2013,Almazan2014} have demonstrated great success in difficult tasks like word spotting or text based retrieval, however they are usually based in expensive techniques. In this scenario the underlying process is similar to the one in multiclass object recognition. 
 It is therefore suggestive for the use of Object Proposals techniques mimicking the state of the art object recognition pipelines.

Traditionally, high precision specialized detectors have been used for segmentation of text in natural scenes, and afterwards text recognition techniques applied to their output~\cite{Gomez2014b}. But it is a well known fact that the perfect text detector, able to work in any conditions, does not exist up to date. In fact, to mitigate the lack of a perfect detector Bissacco~\etal~\cite{bissacco2013} propose an end-to-end scene text recognition pipeline using a combination of several detection methods running in parallel. Demonstrating that if you have a robust recognition method at the end of the pipeline the most important thing in earlier stages is to achieve high recall while precision is not so critical.

The dilemma is thus to choose between having a small set of detections with very high precision but most likely losing some of the words in the scene, or a larger set of proposals, usually in the range of few thousands, with better coverage and then let the recognizer to make the final decision. The later seems to be a well-grounded procedure in the case of word-spotting and retrieval for various reasons. First, as said before, we have powerful whole-word recognizers but they are complex and expensive, second, the recall of current text detection methods may limit their accuracy, and third, sliding window can not be considered an efficient option mainly because words do not have a constrained aspect ratio.
 
In this paper we explore the applicability of Object Proposals techniques in scene text understanding, aiming to produce a set of word proposals with high recall in an efficient way. 
We propose a simple text specific selective search strategy, where initial regions in the image are grouped by agglomerative clustering in a hierarchy where each node defines a possible word hypothesis.
 Moreover, we evaluate different state of the art Object Proposals methods in their ability of detecting text words in natural scenes. We compare the proposals obtained with well known class-independent methods with our own method, demonstrating that our proposal is superior in its ability of producing good quality word proposals in an efficient way.

\section{Related Work}
The use of Object Proposals methods to generate candidate class-independent object locations has become a popular trend in computer vision in recent times. 
A comprehensive survey can be found in Hosang \etal~\cite{hosang2014}. In general terms, we can distinguish between two major types of Object proposals methods: the ones that make use of exhaustive search to evaluate a fast to compute objectness measure~\cite{alexe2012,cheng2014,zitnick2014}, and the ones where the search is segmentation-driven~\cite{Uijlings2013,manen2013,krahenbuhl2014}.

In the first category, Alexe~\etal~\cite{alexe2012} propose a generic objectness measure for a given image window that combines several image cues, such as a saliency score
, the color contrast to its immediate surrounding area, the edge density, and the number of straddling contours. Computation of these features is made efficient by using integral images. 
Cheng~\etal~\cite{cheng2014} propose a very fast objectness score using the norm of image gradients in a sliding window, with a suitable resizing of windows into a small fixed size. A different sliding window driven approach is given by Zitnick~\etal~\cite{zitnick2014}, where a box objectness score is measured as the number of edges~\cite{dollar2013} that are wholly contained in the box minus those that are members of contours that overlap the box's boundary. Using efficient data structures they manage to evaluate millions of candidate boxes in a fraction of second.

On the other hand, selective search methods make use of image's inherent structure through segmentation to guide the search. In this spirit, Gu et al. \cite{Gu2009} make use of a hierarchical segmentation engine \cite{Arbelaez2008} and consider each node in the hierarchy as an object part hypothesis. 
Uijlings et al. \cite{Uijlings2013} argue that a single segmentation and grouping strategy is not enough to generate high quality object locations in any conditions, and thus propose a selective search algorithm that uses multiple complementary strategies. In particular, they start from superpixels using different parameter settings \cite{Felzenszwalb2004} for a variety of color spaces, and then produce a set of hierarchies by merging adjacent regions using different complementary similarity measures. 
Another method based on superpixels merging is due to Manen~\etal\cite{manen2013}, using the connectivity graph induced by the segmentation \cite{Felzenszwalb2004} of an image, with edge weights representing the likelihood that two neighboring pixels belong to the same object, 
their Randomized Prim's algorithm generate proposals by sampling random partial spanning trees with large expected sum of weights. 
 Finally, Kr\"ahenb\"uhl~\etal~\cite{krahenbuhl2014} compute an oversegmentation of the image using a fast edge detector~\cite{dollar2013} and the Geodesic K-means algorithm \cite{perazzi2012}. Then they identify a small set of seed superpixels, aiming to hit all objects in the image, and 
object proposals are identified as critical level sets of the Geodesic Distance Transforms (SGDT) computed for several foreground and background masks for these seeds.


The use of Object Proposals techniques in scene text understanding has been exploited very recently in two state-of-the-art word-spotting methods~\cite{Almazan2015,jaderberg2014reading} while in a distinct manner. In our previous work~\cite{Almazan2015} we propose a text specific selective search method adopting a similar strategy to the selective search of Uijlings~\etal~\cite{Uijlings2013} and a holistic word recognition method based on Fisher Vector representations. On the other hand, Jaderberg~\etal~\cite{jaderberg2014reading} opt for the use of a generic Object Proposals algorithm~\cite{zitnick2014} and deep convolutional neural networks for recognition.


The method proposed in this paper builds on top of our previous work~\cite{Gomez2013,Gomez2014,Almazan2015}, where initial regions in the image are grouped by agglomerative clustering, using complementary similarity measures, in hierarchies where each node defines a possible word hypothesis. But differs from it in two main aspects: First, we do not rely in a classifier to make strong decisions to discriminate text groups from not-text groups, second, we do not combine the different cues in any way.

\section{Text Specific Selective Search}
\label{sec:method}
Our method is based on the fact that text, independently of the script in which it is written, emerges always as a group of similar atomic objects. We make use of the perceptual organisation framework presented in~\cite{Gomez2013}, where a set of complementary grouping cues are used in parallel to generate hierarchies in which each node correspond to a text-group hypotheses. Our algorithm is divided in three main steps: segmentation, creation of hypotheses through bottom-up clustering, and ranking. 

In the first step we use the Maximally Stable Extremal Regions (MSER) algorithm~\cite{Matas2004} to obtain the initial segmentation of the image, as it is proven to be an efficient method for detecting text parts~\cite{Neumann2011}.


\subsection{Creation of hypotheses}
\label{sec:grouping}
The grouping process starts with a set of regions $\mathcal{R}_c$ extracted with the MSER algorithm. Initially each region $r\in\mathcal{R}_c$ starts in its own cluster and then the closest pair of clusters ($A,B$) is merged iteratively, using the single linkage criterion (SLC) ($ \min \, \{\, \mathrm{d}(r_a,r_b) : r_a \in A,\, r_b \in B \,\} $), until all regions are clustered together ($C \equiv \mathcal{R}_c$). Where $\mathrm{d}(r_a,r_b)$ is a distance metric that will be explained next.

Similarly to~\cite{Uijlings2013} we assume that there is no single grouping strategy that is guaranteed to work well in all cases. Thus, our basic agglomerative process is extended with several diversification strategies in order to ensure the detection of the highest number of text regions in any case. First, we extract regions separately from different color channels (i.e. Red, Green, Blue, and Gray) and spatial pyramid levels. Second, on each of the obtained segmentations we apply SLC using different complementary distance metrics:

\begin{equation}
\label{eq:dist}
 \mathrm{d}^{(i)}(r_a,r_b) = (f^{i}(r_a) - f^{i}(r_b))^2 + (x_a - x_b)^2 + (y_a - y_b)^2
\end{equation}

where the term $\{(x_a - x_b)^2 + (y_a - y_b)^2\}$ is the squared Euclidean distance between the centers of regions $r_a$ and $r_b$, and $f(r)$ is a feature aimed to measure the similarity of two regions. Our $f^{i}$ features are designed to exploit the strong similarity of text regions belonging to the same word. We make use of the following simple features with low computation cost: mean gray value of the region, mean gray value in the immediate outer boundary of the region, region's major axis, mean stroke width, and mean of the gradient magnitude at the region's border.

\subsection{Ranking}
\label{sec:rank}
Once we have created our similarity hierarchies each one providing a set of text group hypotheses, we need an efficient way to sort them in order to provide a ranked list of proposals prioritizing the best hypotheses. In the experimental section we explore the use of the following rankings: 

\subsubsection{Pseudo-random ranking}
We make use of the same ranking strategy proposed by Uijlings~\etal in~\cite{Uijlings2013}. Particularly, each hypothesis is assigned with an increasing integer value, starting from 1 for the root node of a hierarchy and subsequently incrementing for the rest of the nodes up to the leaves of the tree. Then each of this values is multiplied with a random number between zero and one, thus providing a ranking that is randomly produced but prioritizes larger regions.
As in~\cite{Uijlings2013} the ranking process is performed before removing duplicate hypotheses. This way if a particular grouping has been found several times within the different hierarchies, indicating a more consistent hypothesis under different similarity cues, this group is going to have more probabilities to be ranked in the top of the list.

\subsubsection{Cluster meaningfulness ranking}
Instead of assigning an increasing value prioritizing larger groups, we propose here to use a cluster quality measure, based on the principle of non-accidentalness, that has been proposed in~\cite{Cao2004} for hierarchical clustering validity assessment. In our case, given one of the grouping cues described in section~\ref{sec:grouping}, equation~\ref{eq:dist} defines a feature space in which individual regions are projected, and the meaningfulness of a group of regions $G$ can be calculated as the inverse of the probability of such a group being a realization of the uniform random distribution:

\begin{equation}
  \mathcal{NFA}(G) = \mathcal{B}_G(k,n,p) = \sum_{i=k}^{n} \binom{n}{i} p^i (1-p)^{n-i}
\label{eq:nfa}
\end{equation}

where $k$ is the number of regions in $G$, $n$ is the total number of regions extracted from the image, and $p$ is the ratio of the volume defined by the distribution of the feature vectors of the regions in $G$ with respect to the total volume of the feature space. Intuitively this value is going to very small for groups comprising a set of very similar regions, that are densely concentrated in small volumes of the feature space. This measure is thus well indicated in the case of measuring text-likeliness of groups because such a strong similarity property is expected to be found in text groups. However, the ranking provided by calculating~\ref{eq:nfa} in each node of our hierarchies is going to prioritize large text groups, e.g. paragraphs, rather that individual words, and thus we combine the ranking provided by equation~\ref{eq:nfa} with a random number between zero and one as done before, providing a pseudo-random ranking where more meaningful hypothesis are prioritized.

\subsubsection{Text classifier confidence}
Finally, we propose the use of a weak classifier to generate our ranking. The basic idea here is to train a classifier to discriminate between text and non-text hypotheses and to produce a confidence value that can be used to rank group hypotheses. Since the classifier is going to be evaluated on every node of our hierarchies, we aim to use a fast classifier and features with low computational cost. We train a Real AdaBoost classifier with decision stumps using as features the coefficients of variation of the individual region features $f^{i}$ described in section~\ref{sec:grouping}: $F^{i}(G) =  {\sigma^{i}}/{\mu^{i}}$, where $\mu^{i}$ and $\sigma^{i}$ are respectively the mean and standard deviation of the region features $f^{i}$ in a particular group $G$,  $\{f^{i}(r) : r \in G\}$. Intuitively the value of $F^{i}$ is smaller for text hypotheses than for non-text groups, and thus the classifier would be able to generate a ranking prioritizing the best hypotheses. 
Notice that all $F^{i}$ group features can be computed efficiently in an incremental way along the SLC hierarchies, and that all $f^{i}$ region features have been previously computed. 



\section{Experiments and Results}

In our experiments we make use of two standard scene text datasets: the ICDAR Robust Reading Competition dataset (ICDAR2013)~\cite{karatzas2013icdar} and the Street View Dataset (SVT)~\cite{Wang2010}. In both cases we provide results for their test sets, consisting in 233 and 249 images respectively, using the original word level ground-truth annotations. 

The evaluation framework used is the standard for Object Proposals methods~\cite{hosang2014} and is based on the analysis of the detection recall achieved by a given method under certain conditions. Recall is calculated as the ratio of GT bounding boxes that have been predicted among the object proposals with an intersection over union (IoU) larger than a given threshold. This way, we evaluate the recall as a function of the number of proposals, and the quality of the first ranked $N$ proposals by calculating their recall at different IoU thresholds. 


\subsection{Evaluation of diversification strategies} 
First, we analyse the performance of different variants of our method by evaluating the combination of diversification strategies presented in Section~\ref{sec:method}. Table~\ref{tab:diversity_comparison} shows the average number of proposals per image, recall rates, and time performance obtained with some of the possible combinations. We select two of them, that we will call ``FAST'' and ``FULL'' as they represent a trade-off between recall and time complexity, for further evaluation.

\begin{table}[h!]
\begin{center}
    \begin{tabular}{ | l | c | c | c | c | c |}
    \hline
    Method & \# prop. & 0.5 IoU & 0.7 IoU & 0.9 IoU & time(s)\\ \hline \hline
    I+D & 536 & 0.84 & 0.65 & 0.41 & 0.26\\ \hline 
    I+DF & 993 & 0.91 & 0.78 & 0.53 & 0.29\\ \hline 
    I+DFBGS & 1323 & 0.95 & 0.86 & 0.60 & 0.45\\ \hline 
    \cellcolor{blue!25}RGB+DF & \cellcolor{blue!25}3359 & \cellcolor{blue!25}0.96 & \cellcolor{blue!25}0.91 & \cellcolor{blue!25}0.69 & \cellcolor{blue!25}0.73\\ \hline 
    RGBI+DFBGS & 5659 & 0.98 & 0.94 & 0.75 & 1.72\\ \hline 
    \cellcolor{blue!25}P2+RGBI+DFBGS & \cellcolor{blue!25}8164 & \cellcolor{blue!25}0.98 & \cellcolor{blue!25}0.96 & \cellcolor{blue!25}0.79 & \cellcolor{blue!25}2.18\\ \hline 
    \end{tabular}
\end{center}
\caption{Max recall at different IoU thresholds and running time comparison of different diversification strategies in the ICDAR2013 dataset. We indicate the use of individual color channels: (R), (G), (B), and (I);  spatial pyramid levels: (P2); and similarity cues: (D) Diameter, (F) Foreground intensity, (B) Background intensity, (G) Gradient, and (S) Stroke width.}
\label{tab:diversity_comparison}
\end{table}

\subsection{Evaluation of proposals' rankings}

\begin{figure}[h!]
\begin{center}
   \includegraphics[width=0.70\linewidth]{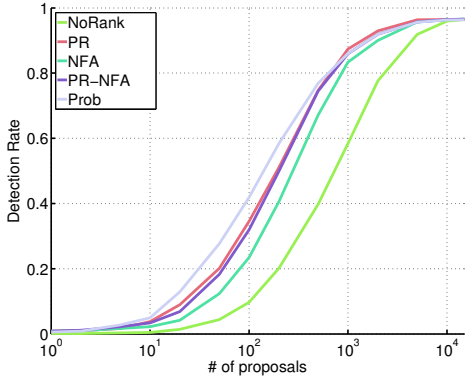}
\end{center}
   \caption{Performance of our ``FAST'' pipeline at 0.5 IoU using different ranking strategies: (PR) Pseudo-random ranking, (NFA) Meaningfulness ranking, (PR-NFA) Randomized NFA ranking, (Prob) the ranking provided by the weak classifier.}
\label{fig:rank_comparison}
\end{figure}

Figure~\ref{fig:rank_comparison} shows the performance of our ``FAST'' pipeline at 0.5 IoU using the various ranking strategies discussed in Section~\ref{sec:method}. The area under the curve (AUC) is 0.39 for NFA, 0.43 both for PR and PR-NFA rankings, while a slightly better 0.46 for the ranking provided by the weak classifier. Since the overhead of using the classifier is negligible we use this ranking strategy for the rest of the experiments.
 
\subsection{Comparison with state of the art} 

\begin{figure*}[t]
\begin{center}
\includegraphics[width=0.32\linewidth]{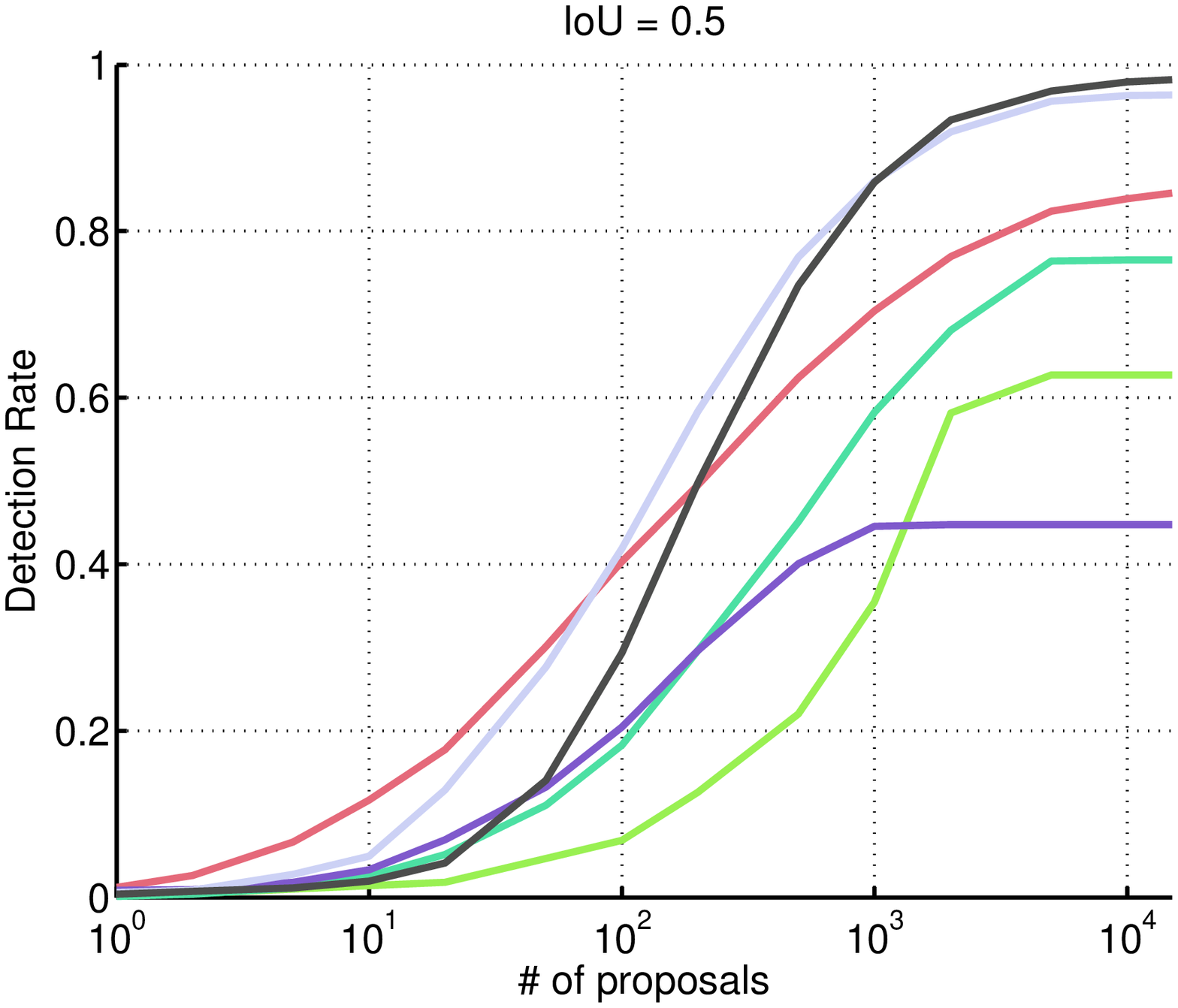} \includegraphics[width=0.32\linewidth]{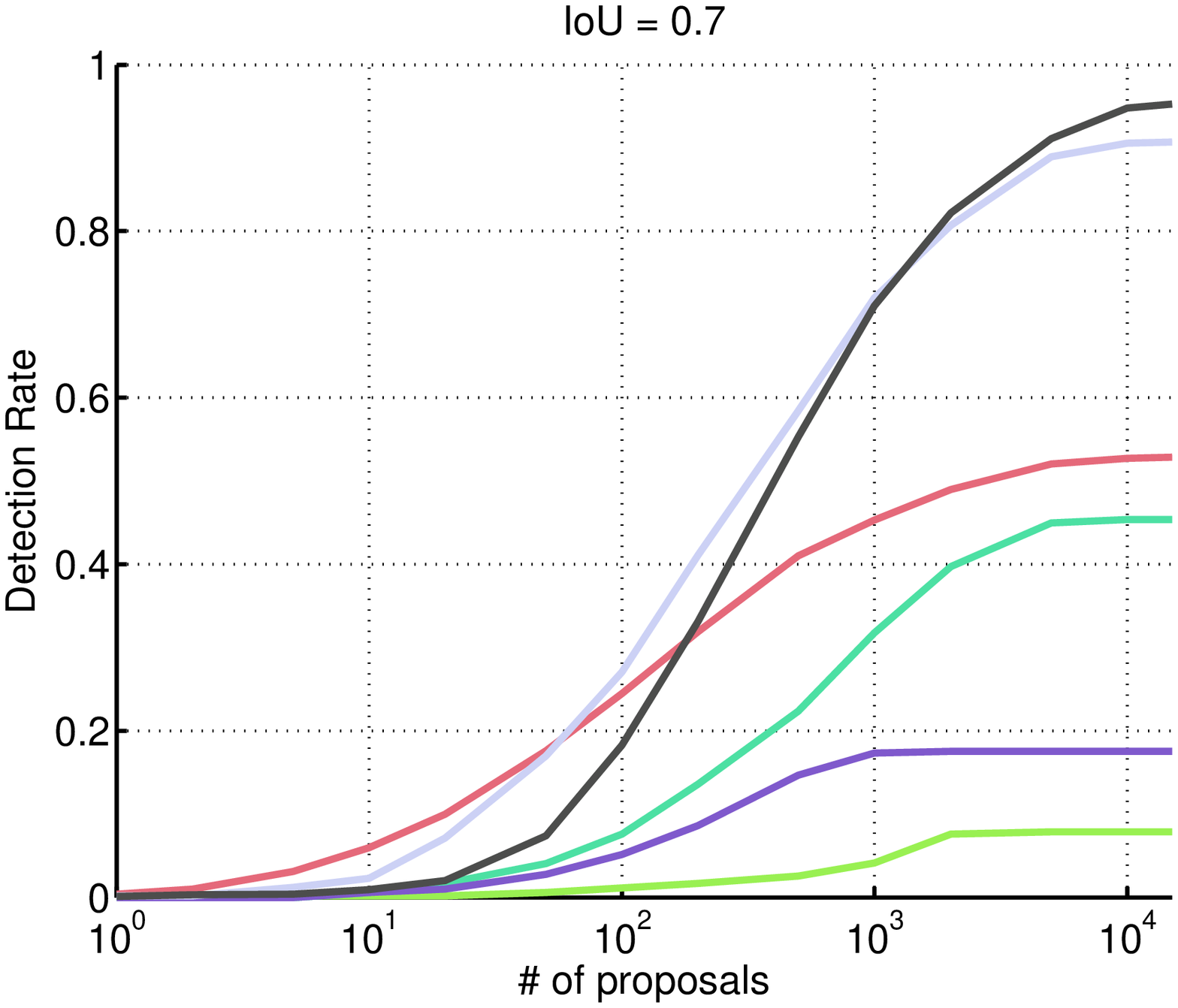} \includegraphics[width=0.32\linewidth]{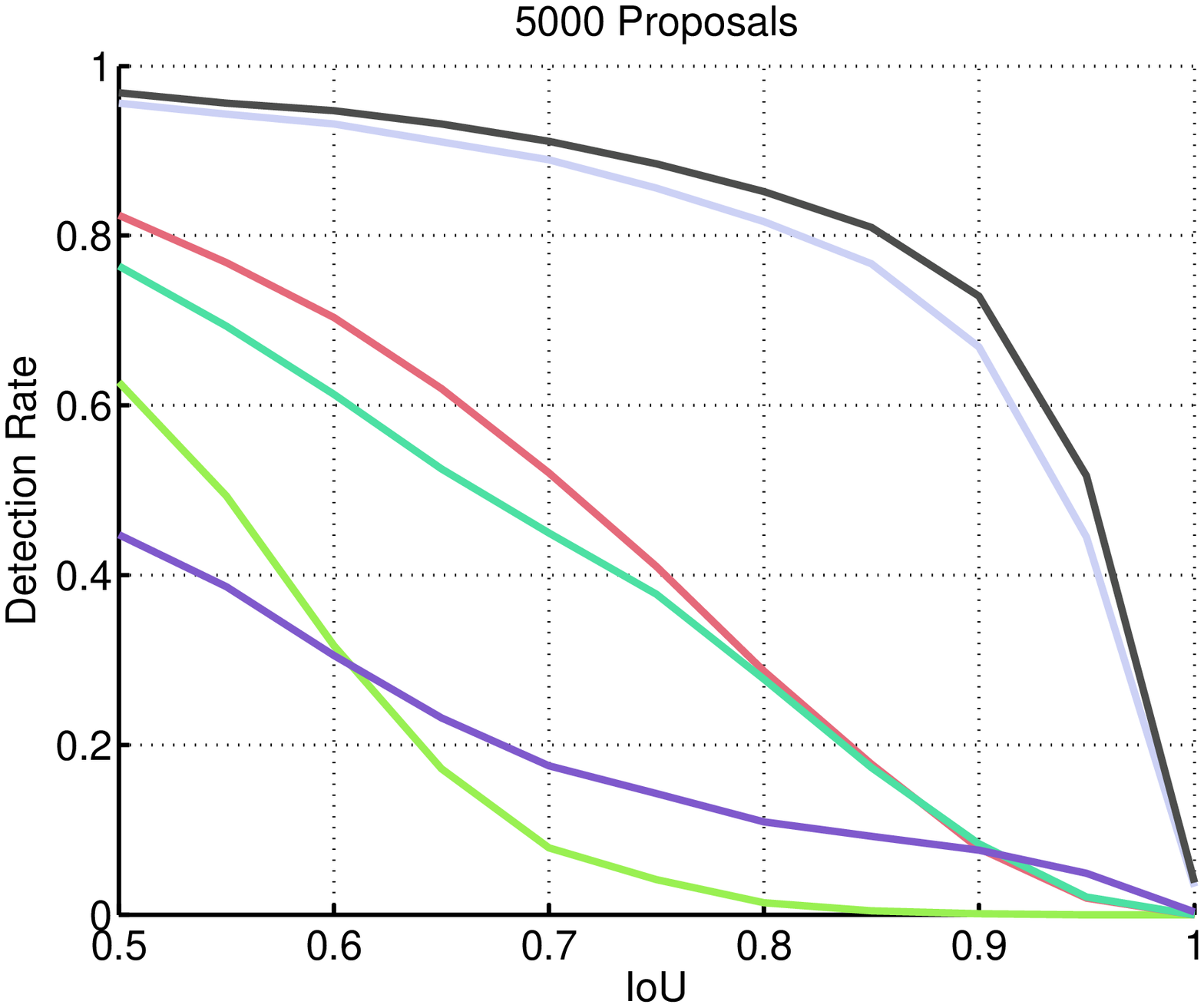}\\
\includegraphics[width=0.90\linewidth]{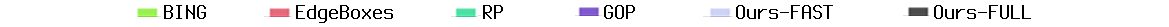}\\
\includegraphics[width=0.32\linewidth]{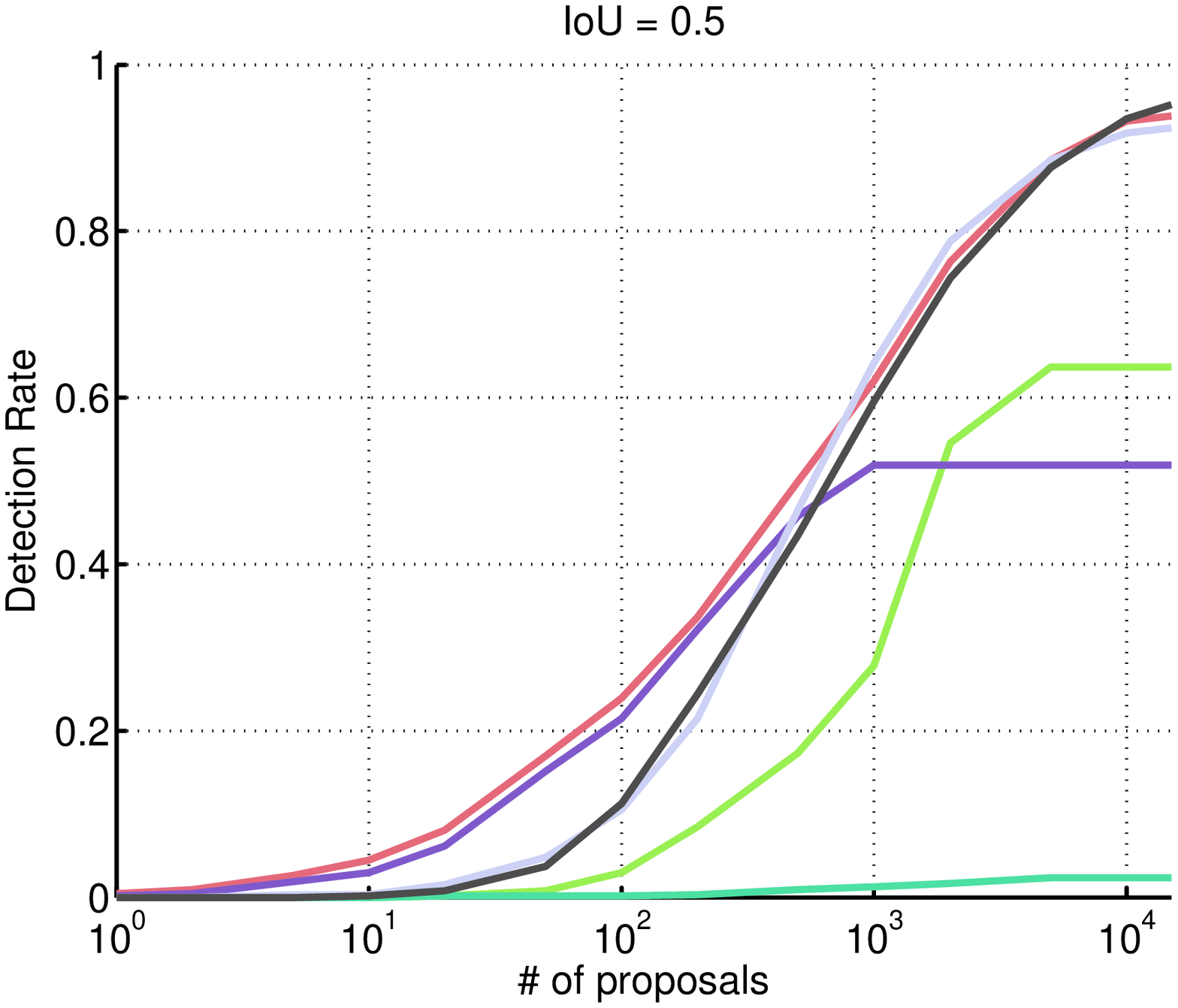} \includegraphics[width=0.32\linewidth]{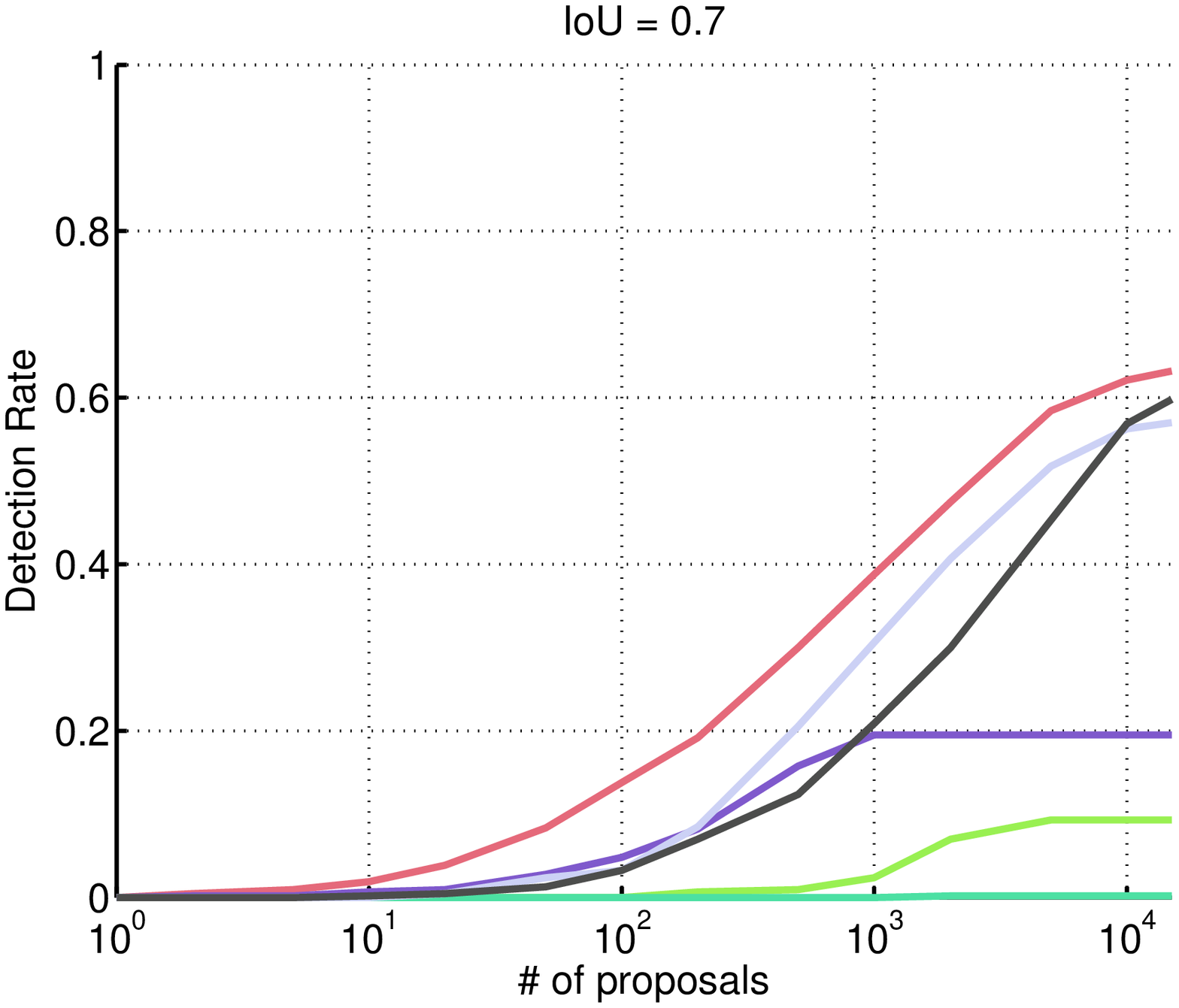} \includegraphics[width=0.32\linewidth]{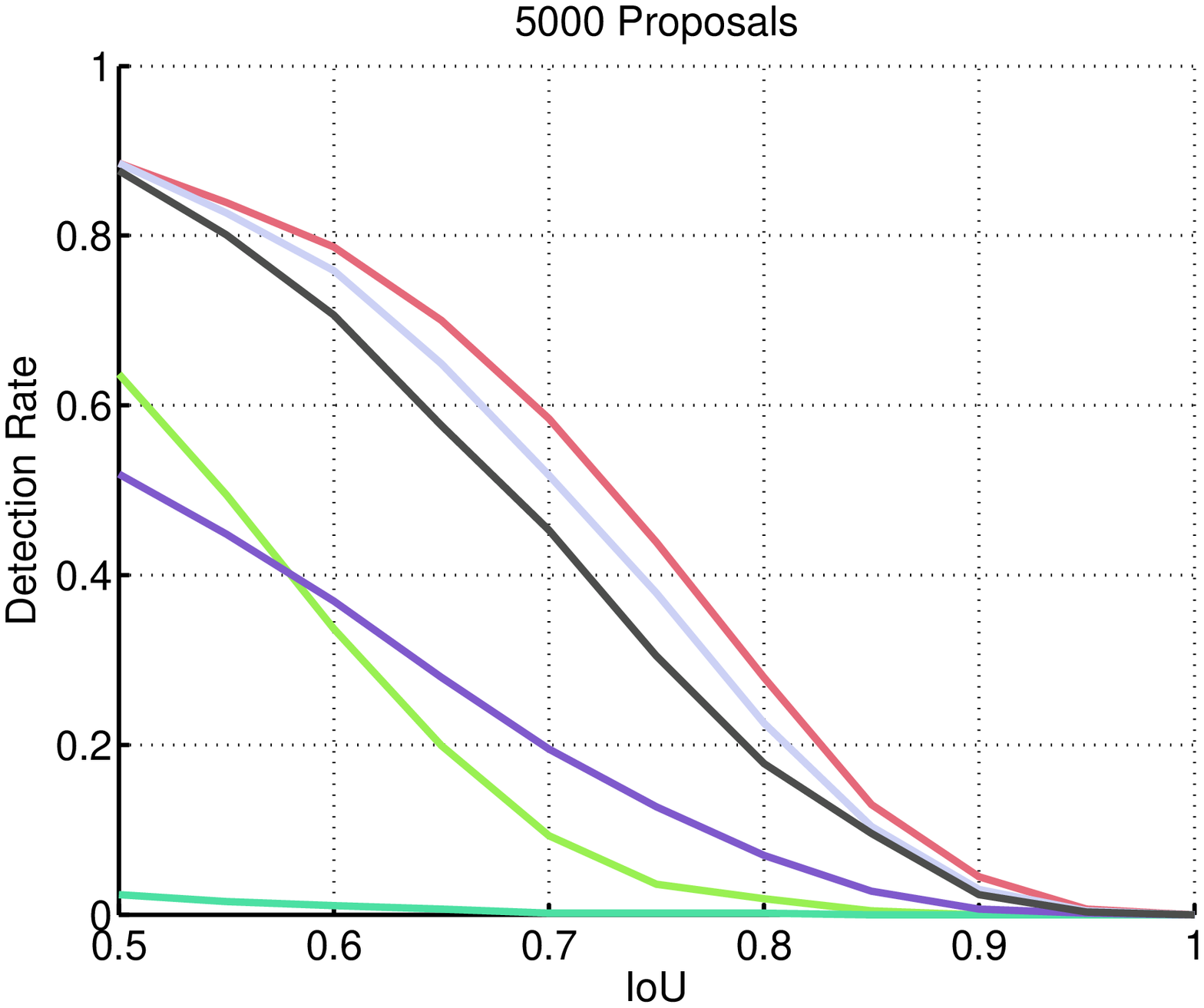}
\end{center}
   \caption{A comparison of various state-of-the-art object proposals methods in the ICDAR2013 (top) and SVT (bottom) datasets. (left and center) Detection rate versus number of proposals for various intersection over union thresholds. (right) Detection rate versus intersection over union threshold for various fixed numbers of proposals.}
\label{fig:plot_icdar}
\end{figure*}

In the following we further evaluate the performance of our method in the ICDAR2013 and SVT datasets, and compare it with the following state of the art Object Proposals methods: BING~\cite{cheng2014}, EdgeBoxes~\cite{zitnick2014}, Randomized Prim's~\cite{manen2013} (RP), and Geodesic Object Proposals~\cite{krahenbuhl2014} (GOP).

In our experiments we use publicly available code of these methods with the following setup. For BING we use the default parameters: base of $2$ for the window size quantization, feature window size of $8\times8$, and non maximal suppression (NMS) size of $2$. For EdgeBoxes we also use the default parameters: step size of the sliding window of $0.65$, and NMS threshold of $0.75$; but we change the max number of boxes to $10^{6}$. GOP is configured with Multi-Scale Structured Forests for the segmentation, $150$ seeds heuristically placed, and $8$ segmentations per seed; in this case we tried other configurations in order to increase the number and quality of the proposals without success. For RP we use the default configuration with $4$ color spaces (HSV,Lab,Opponent,RG) because it provided much better results than sampling from a single graph, while being 4 times slower.

Tables~\ref{tab:icdar_results} and ~\ref{tab:svt_results} show the performance comparison of all the evaluated methods in ICDAR2013 and SVT datasets respectively. A more detailed comparison is provided in Figure~\ref{fig:plot_icdar}. All time measurements in Tables~\ref{tab:icdar_results} and ~\ref{tab:svt_results} have been calculated by executing code in a single thread on the same i7 CPU for fair comparison, while most of them allow parallelization. For instance the multi-threaded version of our method is able to achieve execution times of 0.31 and 0.71 seconds respectively for the ``FAST'' and ``FULL'' variants in the ICDAR2013 dataset.


\begin{table}[h!]
\begin{center}
    \begin{tabular}{ | l | c | c | c | c | c |}
    \hline
    Method & \# prop. & 0.5 IoU & 0.7 IoU & 0.9 IoU & time(s)\\ \hline \hline
    BING~\cite{cheng2014} & 2716 & 0.63 & 0.08 & 0.00 & 1.21\\ \hline 
    EdgeBoxes~\cite{zitnick2014} & 9554 & 0.85 & 0.53 & 0.08 & 2.24\\ \hline 
    RP~\cite{manen2013} & 3393 & 0.77 & 0.45 & 0.08 & 12.80\\ \hline 
    GOP~\cite{krahenbuhl2014} & 855 & 0.45 & 0.18 & 0.08 & 4.76\\ \hline 
    \cellcolor{blue!25}Ours-FAST & \cellcolor{blue!25}3359 & \cellcolor{blue!25}\textbf{0.96} & \cellcolor{blue!25}\textbf{0.91} & \cellcolor{blue!25}\textbf{0.69} & \cellcolor{blue!25}\textbf{0.79}\\ \hline 
    \cellcolor{blue!25}Ours-FULL & \cellcolor{blue!25}8164 & \cellcolor{blue!25}\textbf{0.98} & \cellcolor{blue!25}\textbf{0.96} & \cellcolor{blue!25}\textbf{0.79} & \cellcolor{blue!25}2.25\\ \hline 
    \end{tabular}
\end{center}
\caption{Average number of proposals, recall at different IoU thresholds, and running time comparison with Object Proposals state of the art algorithms in the ICDAR2013 dataset.}
\label{tab:icdar_results}
\end{table}

As can be seen in Table~\ref{tab:icdar_results} and Figure~\ref{fig:plot_icdar} our method outperforms all the evaluated algorithms in terms of detection recall on the ICDAR2013 dataset. Moreover, it is important to notice that detection rates of all the generic Object Proposals heavily deteriorate for large IoU thresholds while our text specific method provides much more stable rates indicating a better coverage of text objects, see the high AUC difference in Figure~\ref{fig:plot_icdar} bottom plots.

\begin{table}[h!]
\begin{center}
    \begin{tabular}{ | l | c | c | c | c | c |}
    \hline
    Method & \# prop. & 0.5 IoU & 0.7 IoU & 0.9 IoU & time(s)\\ \hline \hline
    BING~\cite{cheng2014} & 2987 & 0.64 & 0.09 & 0.00 & 0.81\\ \hline 
    EdgeBoxes~\cite{zitnick2014} & 15319 & 0.94 & \textbf{0.63} & 0.04 & 2.71\\ \hline 
    RP~\cite{manen2013} & 5620 & 0.02 & 0.00 & 0.00 & 10.51\\ \hline 
    GOP~\cite{krahenbuhl2014} & 778 & 0.53 & 0.19 & 0.03 & 4.31\\ \hline 
    \cellcolor{blue!25}Ours-FAST & \cellcolor{blue!25}3791 & \cellcolor{blue!25}0.90 & \cellcolor{blue!25}0.46 & \cellcolor{blue!25}0.03 & \cellcolor{blue!25}\textbf{0.66}\\ \hline 
    \cellcolor{blue!25}Ours-FULL & \cellcolor{blue!25}10365 & \cellcolor{blue!25}\textbf{0.95} & \cellcolor{blue!25}0.61 & \cellcolor{blue!25}\textbf{0.06} & \cellcolor{blue!25}2.22\\ \hline 
    \end{tabular}
\end{center}
\caption{Average number of proposals, recall at different IoU thresholds, and running time comparison with Object Proposals state of the art algorithms in the SVT dataset.}
\label{tab:svt_results}
\end{table}

The results on the SVT dataset in Table\ref{tab:svt_results} and Figure~\ref{fig:plot_icdar} exhibit a radically distinct scenario. While our ``FULL'' pipeline is slightly better than EdgeBoxes at $0.5$ IoU, the later is able to outperform both of our pipelines at $0.7$ and our ``FAST'' variant at $0.5$. Moreover, in this dataset our method does not provide the same stability properties shown before. This can be explained because both datasets are very different in nature, SVT contains more challenging text, with lower quality and many times under bad illumination conditions, while in ICDAR2013 text is mostly well focussed and flatly illuminated. Still, the AUC in most of the plots in Figure~\ref{fig:plot_icdar} show a fairly competitive performance for our method.

\section{Conclusions}
In this paper we have evaluated the performance of generic Object Proposals in the task of detecting text words in natural scenes. We have presented a text specific method that is able to outperform generic methods in many cases, or to show competitive numbers in others. 
Moreover, the proposed algorithm is parameter free and fits well the multi-script and arbitrary oriented text scenario.

An interesting observation of our experiments is that while in class-independent object detection generic methods suffice with near a thousand proposals to achieve high recalls, in the case of text we still need around 10000 in order achieve similar rates, indicating there is a large room for improvement in specific text Object Proposals methods. 

\section{Acknowledgements}
This work was supported by the Spanish project TIN2014-52072-P, the fellowship RYC-2009-05031, and the Catalan govt scholarship 2014FI\_B1-0017.
\small
\bibliographystyle{IEEEtran}
\bibliography{IEEEabrv,GomezKaratzas_icdar15}

\end{document}